\documentclass[letterpaper]{article} 
\usepackage{aaai23}  
\usepackage{times}  
\usepackage{helvet}  
\usepackage{courier}  
\usepackage[hyphens]{url}  
\usepackage{graphicx} 
\usepackage{natbib}  
\usepackage{caption} 
\usepackage{algorithm}
\usepackage{algorithmic}

\usepackage{newfloat}
\usepackage{listings}
\DeclareCaptionStyle{ruled}{labelfont=normalfont,labelsep=colon,strut=off} 
\floatstyle{ruled}
\newfloat{listing}{tb}{lst}{}
\floatname{listing}{Listing}
\usepackage{units}
\usepackage{xcolor}
\usepackage{natbib}
\usepackage{amsmath}
\usepackage{graphicx} 
\usepackage{subfigure}
\usepackage{cleveref}
\usepackage{amsfonts,amssymb}
\usepackage{multirow}
\usepackage[inkscapelatex=false]{svg}

\newtheorem{thm}{Theorem}
\newtheorem{lem}{Lemma}
\newtheorem{remark}{Remark}

\def \hw{\hat{w}}
\def \bw{\tilde{w}}
\def \mw{\mathbf{w}}
\def \mhw{\mathbf{\hat{w}}}
\def \mbw{\mathbf{\tilde{w}}}
\def \md{\mathbf{\Delta}}
\def \CE{\mathbf{E}}
\def \r{{\mathbf{r}}}

\title{Adaptive Low-Precision Training for Embeddings in Click-Through Rate Prediction}
\author{
    Shiwei Li,\textsuperscript{\rm 1}\thanks{This work was done when Shiwei Li worked as an intern at Noah's Ark Lab, Huawei.}
    Huifeng Guo,\textsuperscript{\rm 2}
    Lu Hou,\textsuperscript{\rm 2}
    Wei Zhang,\textsuperscript{\rm 2}
    Xing Tang,\textsuperscript{\rm 2}\\
    Ruiming Tang,\textsuperscript{\rm 2}
    Rui Zhang,\textsuperscript{\rm 3}\footnotemark[2]
    Ruixuan Li\textsuperscript{\rm 1}\thanks{Corresponding authors.}
}
\affiliations {
    \textsuperscript{\rm 1} Huazhong University of Science and Technology, Wuhan, China\\
    \textsuperscript{\rm 2} Huawei Noah's Ark Lab, Shenzhen, China\\
    \textsuperscript{\rm 3} www.ruizhang.info\\
    \{d202181195, rxli\}@hust.edu.cn,
    \{rayteam\}@yeah.net,\\
    \{huifeng.guo, houlu3, xing.tang, tangruiming\}@huawei.com
}

\begin{document}
\maketitle
\begin{abstract}
Embedding tables are usually huge in click-through rate (CTR) prediction models. To train and deploy the CTR models efficiently and economically, it is necessary to compress their embedding tables at the training stage. To this end, we formulate a novel quantization training paradigm to compress the embeddings from the training stage, termed low-precision training (LPT). Also, we provide theoretical analysis on its convergence. The results show that stochastic weight quantization has a faster convergence rate and a smaller convergence error than deterministic weight quantization in LPT. Further, to reduce the accuracy degradation, we propose adaptive low-precision training (ALPT) that learns the step size (i.e., the quantization resolution) through gradient descent. Experiments on two real-world datasets confirm our analysis and show that ALPT can significantly improve the prediction accuracy, especially at extremely low bit widths. For the first time in CTR models, we successfully train 8-bit embeddings without sacrificing prediction accuracy. The code of ALPT is publicly available\footnote{\url{https://gitee.com/mindspore/models/tree/master/research/recommend/ALPT}}.
\end{abstract}

\section{Introduction}\label{sec:introduction}
Click-through rate (CTR) prediction is to predict the probability that a user will click on a recommended item under a specific context~\cite{wide_deep}, which is a critical component in recommender systems. It is widely used in various scenarios, such as online shopping~\cite{din} and advertising~\cite{ftrl}. 
With the development of deep neural networks, CTR models evolve from logistic regression~\cite{ftrl}, factorization machine models~\cite{ffm} to deep learning models. 
Various deep CTR models have been proposed and deployed in industrial companies, such as Wide \& Deep~\cite{wide_deep} in Google, DIN~\cite{din} in Alibaba and and DeepFM~\cite{DeepFM} in Huawei.

\begin{figure}[h]
\centering
\includegraphics[scale=0.56] {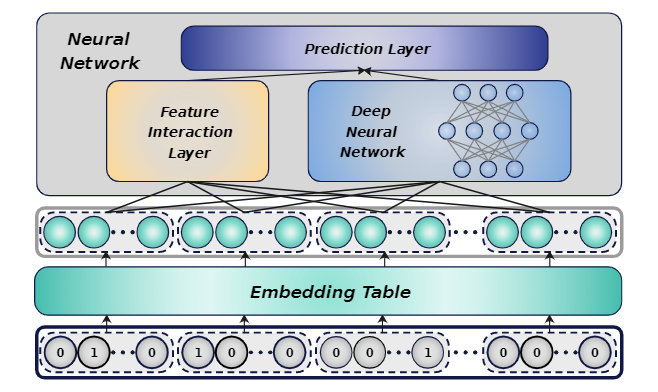}
\caption{The \textit{embedding table} and \textit{neural network} paradigm of click-through rate prediction models.}\label{fig:embnn}
\end{figure}

Deep CTR models usually follow the \textit{embedding table} and \textit{neural network} paradigm~\cite{autodis}. 
As shown in Figure~\ref{fig:embnn}, the \textit{embedding table} transforms the one-hot encoded vector of high-dimensional categorical features, such as \texttt{user\_id} and \texttt{item\_id}, into low-dimensional dense real-valued vectors (i.e., embeddings)~\cite{DeepFM}. 
The \textit{neural network} is used to model the feature interactions and then make predictions. 
Usually, each feature has a unique embedding stored in the embedding table $\mathbf{E}\in \mathbb{R}^{n\times d}$, where $n$ is the number of features and $d$ is the embedding dimension.  However, since there are usually billions or even trillions of categorical features, the embedding tables may take hundreds of GB or even TB to hold~\cite{sfctr}. 
For example, the size of embedding tables in Baidu's advertising systems reaches 10 TB~\cite{baidu_int16}. 

To deploy the CTR models with huge embedding tables in real production systems efficiently and economically, it is necessary to compress their embedding tables. 
Research on embedding compression in recommender systems has mostly focused on three aspects: 
(i) NAS-based embedding dimension search~\cite{NIS, AutoDim}; 
(ii) Embedding pruning~\cite{DeepLight, PEP}; 
(iii) Hashing~\cite{QRhash, doublehash}. 
Unfortunately, these approaches are usually not practical in real recommender systems. 
Specifically, NAS-based approaches require additional storage and complex calculations to search for the optimal embedding dimension. Embedding pruning approaches usually produce unstructured embedding tables which will cost extra effort to access. 
Note that the models trained with these two approaches should be retrained to maintain accuracy. 
Besides, they can not compress the embeddings at the training stage, which although hash-based approaches can do, they usually cause severe degradation in prediction accuracy~\cite{baidu_int16,QRhash}. 
In industrial recommender systems, the CTR models with extremely large embedding tables usually require distributed training on multiple devices~\cite{baidu_int16}, where the communication between multiple devices seriously affects the training efficiency. By compressing the embeddings at training stages, CTR models can be trained on less devices or even one single GPU, which can accelerate training by reducing the communication cost and taking full advantage of the computing power of the hardware.

In this paper, we aim to compress the large embedding tables in CTR models from the training stage without sacrificing accuracy by quantization. As far as we know, the quantization schemes in existing works~\cite{baidu_int16,cache-facebook} are also proposed for the same purpose. We term this quantization scheme as low-precision training (LPT). 
Specifically, LPT keeps the weights in low-precision format during training. Different from the commonly-used quantization-aware training (QAT)~\cite{courbariaux2015binaryconnect,lsq-EsserMBAM20} which keeps a copy of the full-precision weights for parameter update, LPT directly updates low-precision weights and then quantizes the updated full-precision weights back into low-precision format. 

However, without the copy of full-precision weights, LPT usually suffers from inferior performance than QAT.
\cite{baidu_int16} claims that their maximum ability is using 16-bit LPT for embeddings, since lower bit-widths result in unacceptable accuracy degradation. Although \cite{cache-facebook} achieves lossless 8-bit LPT for  embeddings, they have to keep a full-precision cache which brings extra memory cost. 
The efforts of \cite{baidu_int16} and \cite{cache-facebook} illustrate that training low-precision embeddings is very challenging, especially for lower bit-widths. However, they are only empirical and lack theoretical analysis for LPT. 
Besides, they did not explore the impact of the step size (i.e., the quantization resolution) on the accuracy.  
Therefore, to explore the limitations of LPT, we first provide theoretical analysis on its convergence. Further, to reduce the accuracy loss, we propose adaptive low-precision training to learn the step size. The contribution are summarized  as follows:

(1) We formulate a low-precision training paradigm for compressing embedding tables at the training stage and provide theoretical analysis on its convergence. The results show that stochastic weight quantization has a faster convergence rate and a smaller convergence error than deterministic weight quantization in LPT.

(2) Different from previous works, we offer a solution to learn the step size with gradient descent in LPT, termed adaptive low-precision training (ALPT). 

(3) Experiments are conducted on two public real-world datasets for CTR prediction tasks. The results confirm our theoretical analysis and show that ALPT can significantly improve the prediction accuracy. Moreover, ALPT can train 8-bit embeddings without sacrificing prediction accuracy, which is achieved the first time, ever.

\section{Preliminaries}\label{sec:preliminaries}
In this section, we elaborate on the training process of LPT. 
In Section~\ref{sec:quantization}, we first introduce how quantization works. 
In Section~\ref{sec:qat}, we introduce the commonly-used QAT, as it is important to understand how LPT differs from it. In Section~\ref{sec:lpt}, we introduce LPT and explain why it can compress the embedding tables of CTR models at the training stage.

\subsection{Quantization} \label{sec:quantization}
Quantization {compresses a network by replacing the 32-bit full-precision weights with their  lower-bit counterparts without changing the network architecture.}
Specifically, for $m$-bit quantization, the set of the quantized values can be denoted as $\mathbb{S}=\{ b_0, b_1, ...,b_k \}$, where $k=2^m-1$.
One commonly used quantization scheme is the uniform quantization, where the quantized values are uniformly distributed, and is usually more hardware-friendly than the non-uniform quantization. 
In the uniform symmetric quantization, the step size $\Delta = b_{i}-b_{i-1}$ remains the same for any $i \in [1,k]$, and $b_0=-2^{m-1}\Delta, b_k=(2^{m-1}-1)\Delta$.

In this paper, we adopt uniform symmetric quantization on the embeddings. Specifically, given the step size $\Delta$ and the bit width $m$, a full-precision weight $w$ is quantized into $\hw \in \mathbb{S}$, which is represented by the multiplication of the step size $\Delta$ and an integer $\bw$:

\begin{equation}\label{eq:1}
\bw = {R}({clip}(w/\Delta,-2^{m-1},2^{m-1}-1)),
\end{equation}
\begin{equation}\label{eq:2}
\hw = Q(w)= \Delta {\times} \bw,
\end{equation}
where ${clip}(v,n,p)$ returns $v$ with values below $n$ set to $n$ and values above $p$ set to $p$ and $ R(v)$ rounds $v$ to an adjacent integer. 
There are generally two kinds of rounding functions: 

\textbf{Deterministic Rounding (DR)} 
rounds a floating-point value to its nearest integer:
\begin{equation}
R_D(x) = \left\{
\begin{array}{ll}
    \lfloor x\rfloor    & \text{if} \ x-\lfloor x\rfloor < 0.5,\\ 
    \lfloor x\rfloor+1  & \text{otherwise}.
\end{array}
\right. \label{eq:stoc}
\end{equation}

\textbf{Stochastic Rounding (SR)} 
rounds a floating-point value to its two adjacent integers with a probability distribution:
\begin{equation}
R_S(x) = \left\{
\begin{array}{ll}
    \lfloor x\rfloor    & \text{w.p.}\quad \lfloor x\rfloor+1-x,\\ 
    \lfloor x\rfloor+1  & \text{w.p.}\quad x-\lfloor x\rfloor.
\end{array}
\right. \label{eq:stoc}
\end{equation}
To distinguish these two different rounding functions, we add subscripts for the quantization function in Eq.~(\ref{eq:2}), that is $Q_D$ and $Q_S$ for DR and SR, respectively. Note that $\Delta$ is also a parameter of the quantization function $Q(w,\Delta)$, here we omit $\Delta$ for simplicity.

\subsection{Quantization-aware training}\label{sec:qat}
\begin{figure}[t]
\centering
\subfigure[Quantization-aware training] 
{
\begin{minipage}[t]{0.45\textwidth}\label{fig:qat}
\centering
\includegraphics[scale=0.3]{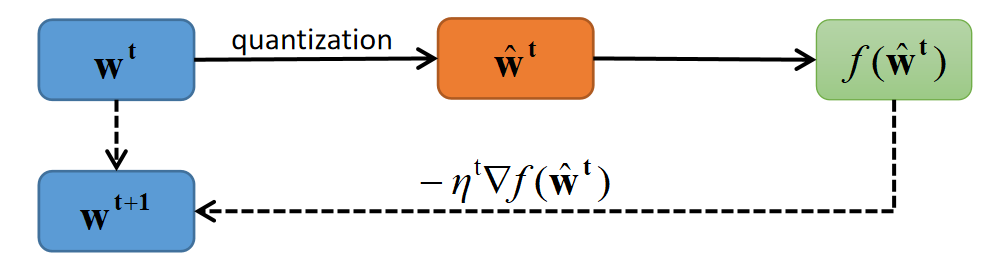}
\end{minipage}
}
\subfigure[Low-precision training] 
{
\begin{minipage}[t]{0.45\textwidth}\label{fig:lpt}
\centering
\includegraphics[scale=0.3]{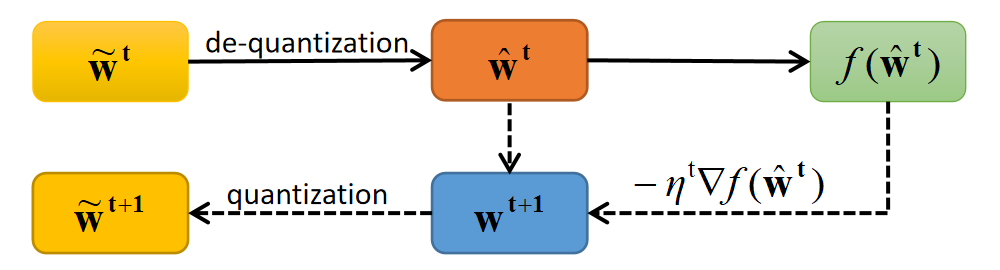}
\end{minipage}
}
\caption{Training processes of quantization-aware training (QAT) and low-precision training (LPT).}
\label{fig:qat_tlp}
\end{figure}

As Figure~\ref{fig:qat} shows, 
quantization-aware training (QAT) quantizes the full-precision weights in the forward pass and updates the full-precision weights with the gradients estimated by straight through estimator (STE) \cite{courbariaux2015binaryconnect}. 
Specifically, let $f(\cdot)$ be the loss function and $\nabla f(\mhw^{t})$ be the gradients w.r.t. the quantized weights $\mhw^{t}$. For stochastic gradient descent with a learning rate of $\eta^t$, the full-precision weights will be updated as:
\begin{equation}\label{eq:qat}
\mw^{t+1} = \mw^{t} - \eta^t  \nabla f(\mhw^{t}).
\end{equation}

To achieve better accuracy, recent methods have offered solutions to learn the step size. 
For example, LSQ~\cite{lsq-EsserMBAM20} uses the following quantization function:
\begin{equation}
Q_D(w^t_i) = \Delta {\times} R_D(\text{Clip}(\frac{w^t_i}{\Delta},-2^{m-1},2^{m-1}-1)),
\end{equation} 
and optimizes $\Delta$ through gradient descent where the gradient of the step size $\Delta$ is estimated by:
\begin{equation}\label{eq:grad}
\frac{ \partial  Q_D(w^t_i) }{ \partial \Delta } = \left\{
\begin{array}{ll}
    -2^{m-1}                & \text{if} \ w^t_i/\Delta \leq  -2^{m-1} ,\\ 
    2^{m-1}-1               & \text{if} \ w^t_i/\Delta \geq 2^{m-1}-1 ,\\ 
    R_D(\frac{w^t_i}{\Delta})-\frac{w^t_i}{\Delta}   & \text{otherwise}.\\
\end{array}
\right.
\end{equation}

After training, each weight matrix can be stored in the format of integers plus one full-precision step size. However, in the training process of QAT, the full-precision weights are involved in the update process, which means QAT can only compress the model size for inference.  

\subsection{Low-precision training}\label{sec:lpt}
As Figure~\ref{fig:lpt} shows, unlike QAT which still keeps a copy of the full-precision weights for parameter update, low-precision training (LPT) directly updates the low-precision weights and then quantize the updated full-precision weights back into low-precision format as:
\begin{equation}\label{eq:lpt}
\mhw^{t+1} = Q(\mhw^{t} - \eta ^t  \nabla f(\mhw^{t})).
\end{equation}

For the embedding tables in CTR models, LPT is quite effective in compression of training memory. 
Note that embedding tables are usually highly sparse and each batch of the training data only covers very few features.
For example, in our processed dataset Avazu which has 24 feature fields and more than 4 million features, a batch of ten thousand samples only contains 1400 features on average.  
With LPT, the whole embedding table can be stored in the format of integers and only the embeddings of very few features that appear in each batch will be de-quantized into floating-point values for calculation and update. The storage of the step size and the de-quantized floating point weights are negligible compared to the embedding tables. 
In this way, LPT can effectively compress the model at the training stage, however, the issue of inferior accuracy remains to be resolved. 
In this paper, we aim to improve the accuracy of LPT from the perspective of the rounding functions and the step size.

\section{Methodology}\label{sec:methodology}
As discussed in Section \ref{sec:lpt}, we need to quantize the updated full-precision weights back into low-precision format in LPT. When choosing a quantization function, we have to consider two key issues: 
 (i) which rounding function suits LPT better? 
 (ii) how to select reasonable step size flexibly and effectively? 
To address the first issue, we first analyze the convergence of SR and DR for LPT in Section~\ref{sec:sr-dr}. To address the second issue, in Section~\ref{sec:alpt}, we first point out the difficulties of learning the step size in LPT, then we introduce our adaptive low-precision training algorithm. 

\subsection{Rounding: stochastic or deterministic?}\label{sec:sr-dr}
Generally, DR is the common choice of quantization as it produces lower mean-square-error (MSE) \cite{lsq-EsserMBAM20,pact}. Still, recent works \cite{baidu_int16,li2017training} argue that SR can achieve better performance in LPT. However, they did not provide theoretical analysis about the difference between SR and DR in LPT. Therefore, in this section, we first provide theoretical analysis on the convergence of SR and DR in LPT, and then we use a synthetic experiment to visualize the difference between SR and DR.  

\subsubsection{Convergence analysis}
To analyze the convergence for LPT, we consider empirical risk minimization problems as in Eq.(\ref{eq:risk}) following \cite{li2017training}. $\mw$ is used to denote all the parameters in a model and the loss function is decomposed into a sum of multiple loss functions $\mathbb{F}=\{f_1,f_2,...,f_m\}$:

\begin{equation}\label{eq:risk}
\min_{\mw\in W}F(\mw) :=\frac{1}{m}\sum _{i=1}^m f_i(\mw).
\end{equation}
 At the $t$-th iteration of the gradient descent, we 
select a function $ f^t \in \mathbb{F}$ and update the model parameters as:
\begin{equation}
\mw^{t+1} = \mw^{t}-\eta^t \nabla  f^t(\mw^{t}).
\end{equation}
\cite{li2017training} already provides convergence analysis for SR in LPT (Theorem~\ref{thm:stoc_conv}). 
However, it lacks the analysis for DR. Since DR is biased (i.e., the expectation of the quantization error is not zero), it is quite challenging to extend the conclusion of Theorem~\ref{thm:stoc_conv} to DR. 
In Theorem~\ref{thm:det_conv}, we  provide convergence analysis for DR in LPT, using the same assumptions as \cite{li2017training}: 
(i)	each $f_i \in \mathbb{F}$ is differentiable and convex;
(ii) $f^t(\mw^t)$ has bounded gradient, i.e., $\CE||\nabla f^t(\mw^t)||^2 \leq G^2$; and
(iii) the domain of $\mw$
is a convex set and has finite diameter, i.e., $||\mw^m - \mw^n)||^2 \leq D^2$.

\begin{thm} \label{thm:stoc_conv} 
$\left[ \text{Theorem 2 in \cite{li2017training}} \right]$ 
Assume the learning rate decays like $\eta ^t=\frac{\eta }{\sqrt{t}}$. At the $T$-th iteration, {with a fixed step size $\Delta$,} for SR in LPT, we have: 
\begin{equation}
\CE\left[F(\bar\mw^T)-F(\mw^*)\right] \leq \frac{D^2}{2\eta \sqrt{T}}
+ \frac{\eta  G^2}{\sqrt{T}}
+ \frac{\sqrt{d}\Delta G}{2},
\end{equation}
\end{thm}
where $\bar\mw^T=\frac{1}{T}\sum_{t=1}^T \mw^t$, $\mw^*={\arg\min}_{\mw} F(\mw)$ and $d$ is the dimension of $\mw$.

\begin{thm} \label{thm:det_conv}
Assume the learning rate decays like $\eta ^t=\frac{\eta }{\sqrt{t}}$. At the $T$-th iteration, with a fixed step size $\Delta$, let $T_0=\lfloor \frac{2\eta G} {\sqrt{d}\Delta} \rfloor$. For DR in LPT, we have: 
\begin{equation}
\begin{aligned}
\CE\left[F(\bar\mw^T)-F(\mw^*)\right] 
&\leq  \frac{D^2}{2\eta \sqrt{T}}
+ \frac{3\eta  G^2}{\sqrt{T}} + \frac{\sqrt{d}\Delta G}{2}  \\
+ &\frac{\sqrt{d}D\Delta\sum_{t=1}^{T_0} \sqrt{t}}{2\eta T} + \frac{\sum_{t=T_0+1}^{T}DG}{T}. 
\end{aligned}
\end{equation}
\end{thm}

The detailed proofs are presented in Appendix~\ref{sec:proof}. The theorems illustrate that both DR and SR converge to an accuracy floor in LPT. However, DR has a slower convergence rate and a larger convergence error than SR, which proves that SR is more suitable for LPT. 

\begin{remark}\label{rmk:1}
If gradient updates are extremely small and $|\eta ^t \nabla f(w_i^{t})| \textless \frac{\Delta}{2}$, DR will erase the update of $w_i^t$ and $w_i$ may never change from $w_i^t$. Such error accumulation results in slower convergence rates and larger convergence errors.
\end{remark}

\begin{remark} 
Here we only provide analysis based on a fixed step size. Empirically, we shall show in Section~\ref{sec:experiment} that using an adaptive step size in Section~\ref{sec:alpt} achieves better accuracy. 
\end{remark}

\subsubsection{Synthetic experiment}
\begin{figure}[h]
\centering
\subfigure[$t=10$]
{
 	\begin{minipage}[b]{.45\linewidth}\label{fig:toy:1}
        \centering
        \includegraphics[scale=0.3]{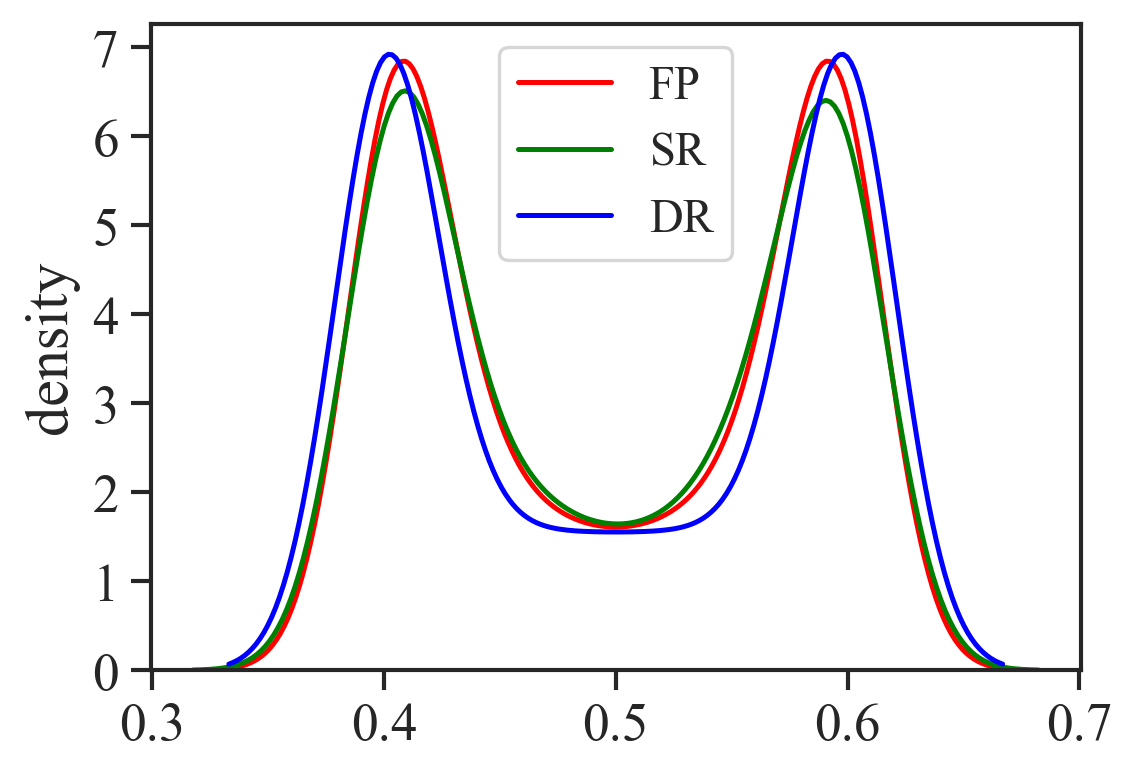}
    \end{minipage}
}
\subfigure[$t=100$]
{
 	\begin{minipage}[b]{.45\linewidth}\label{fig:toy:2}
        \centering
        \includegraphics[scale=0.3]{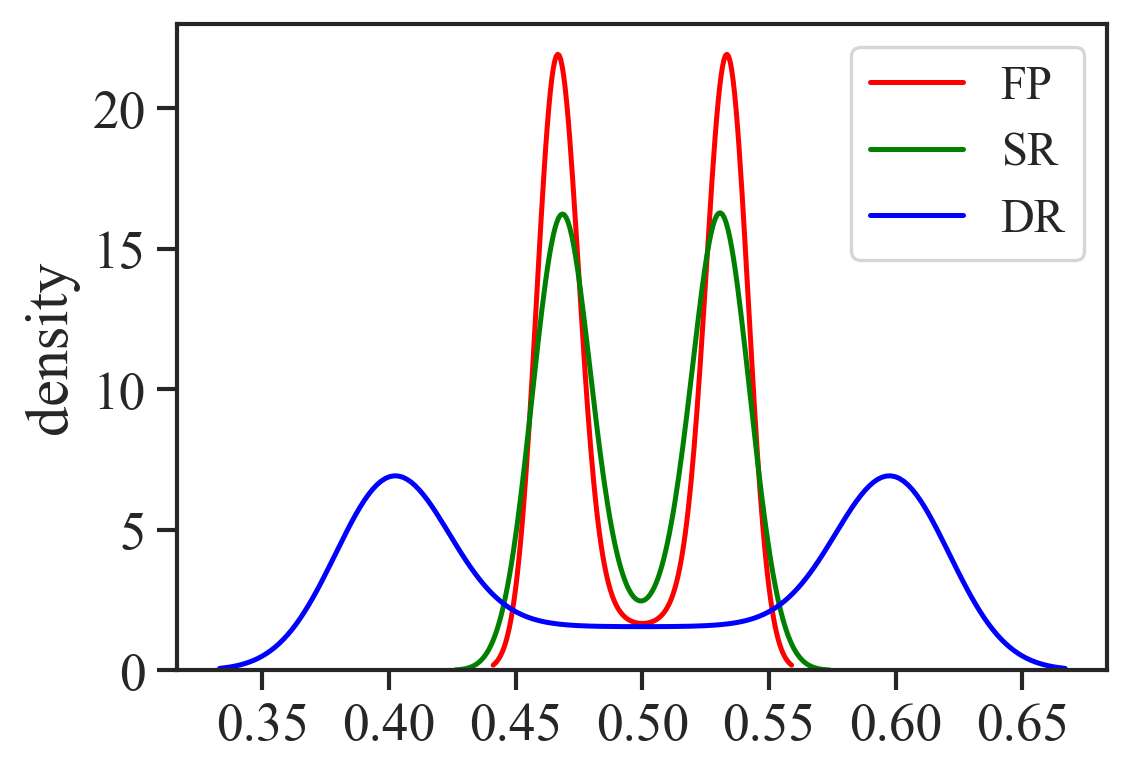}
    \end{minipage}
}
\subfigure[$t=1000$]
{
 	\begin{minipage}[b]{.45\linewidth}\label{fig:toy:3}
        \centering
        \includegraphics[scale=0.3]{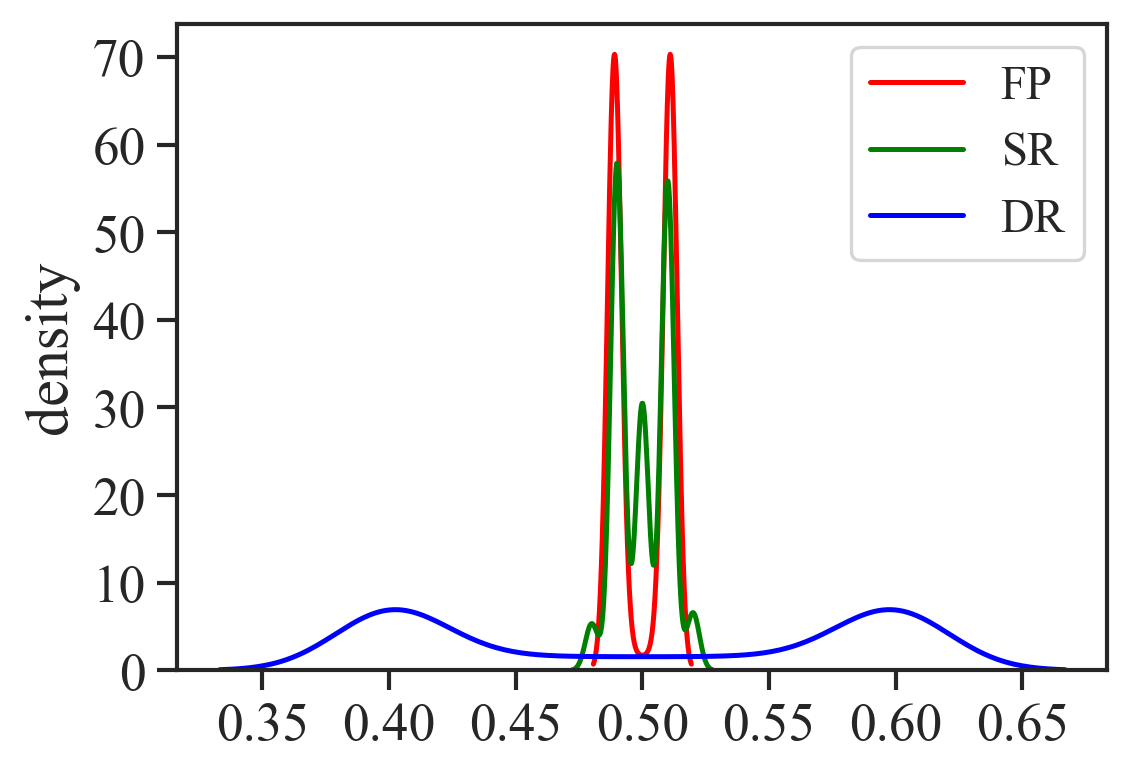}
    \end{minipage}
}
\subfigure[Gradients in DR]
{
 	\begin{minipage}[b]{.45\linewidth}\label{fig:toy:4}
        \centering
        \includegraphics[scale=0.3]{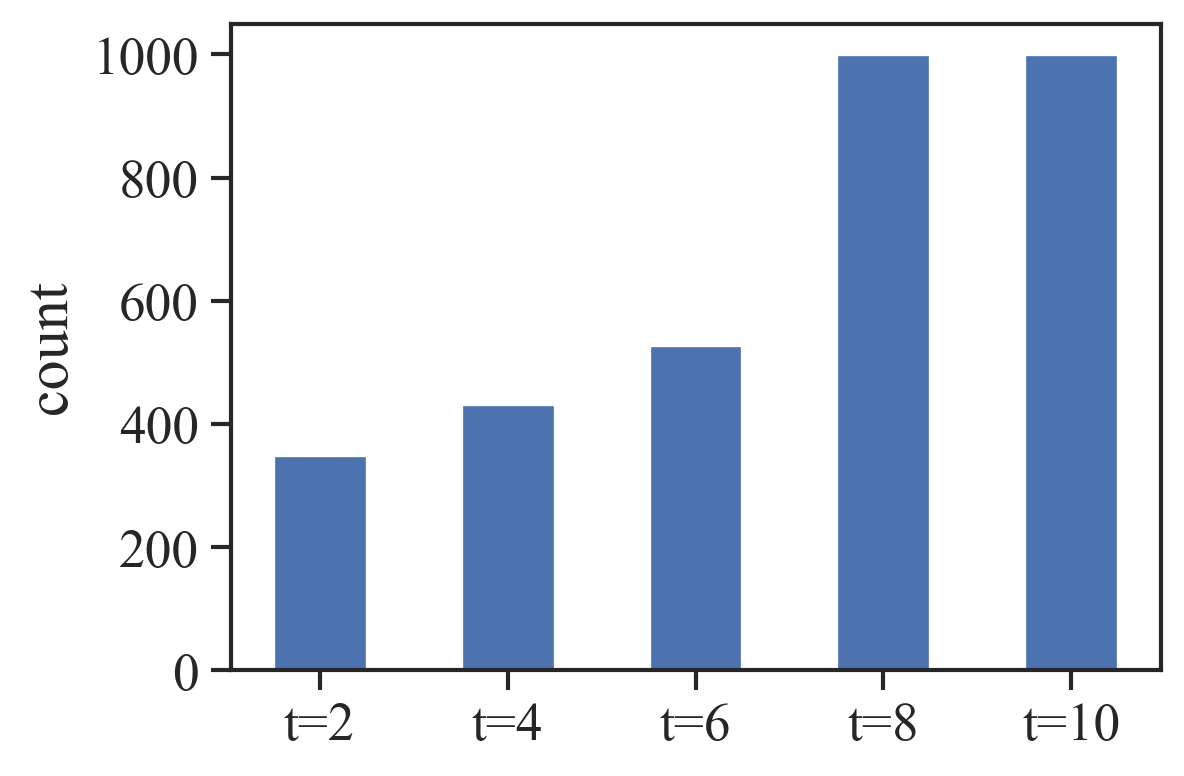}
    \end{minipage}
}
\caption{(a), (b) and (c) plot the distributions of the parameters. FP stands for training with full-precision parameters; DR and SR stands for training with DR and SR in LPT, respectively. (d) plots the number of parameters that satisfy $|\eta ^t \nabla f(w^{t})| \textless \frac{\Delta}{2}$ at different iterations of DR.}
\end{figure}

To visualize the difference of DR and SR in LPT, we design a simple convex problem:
\begin{equation}
\mathop{\min}\limits_{w} {f}(w) = (w-0.5)^2.
\end{equation}
We initialize 1000 parameters between 0 and 1 uniformly. Each parameter is updated by SGD with learning rate $\eta = 1$. We set $\Delta=0.01$ and $m=8$ for quantization. 
Figures~\ref{fig:toy:1}, \ref{fig:toy:2}, \ref{fig:toy:3} show the distribution of the parameters at the iteration $t=$ 10, 100 and 1000, respectively. 
As we can see, SR have a similar or even faster convergence rate compared to the full-precision parameters, while DR seem to be stagnant as described in Remark~\ref{rmk:1}. 
As Figure~\ref{fig:toy:4} shows, in LPT with DR, all the gradients satisfy $|\eta ^t \nabla f(w^{t})| \textless \frac{\Delta}{2}$ after 10 iterations that is the moment when the parameters stop updating.

\subsection{Adaptive low-precision training}\label{sec:alpt}
In the previous section, we have demonstrated that SR is more suitable for LPT, yet another key factor that affects the performance is the step size. 
Intuitively, since the representation range is proportional to the step size given the bit-width, a small step size will lead to a limited representation range. On the contrary, a large step size can not provide fine resolution to the majority of weights within the clipping range. Therefore, it is important to select a reasonable step size.
To figure out how to find the desired step size, we first consider two preliminary problems: 
\begin{itemize}
    \item \textbf{Manually or adaptively?} If we set the step size as a hyper-parameter, it would take a lot of human efforts to tune the step size for different applications. Nevertheless, the tuned result may still not be optimal. Thus, 
we should make the step size adaptive that is to learn the step size together with the embeddings in an end-to-end manner. 

\item \textbf{Global-wise or feature-wise?} A global step size will be shared throughout the whole embedding table, which will fail to trade off the representation range and precision for all embeddings. Besides, when a global step size updates, all the embeddings are supposed to be re-quantized with the updated step size, which significantly reduce the training efficiency. On the contrary, a feature-wise step size is bound with its owner feature and is free to  update.
\end{itemize}

In light of the above discussion, we should learn a step size for each embedding. 
However, there is no gradients for the step size in LPT as the quantization is taken place after the backward propagation. 
To learn the step size, we shall introduce the step size to the forward propagation.
However, even if we quantize the low-precision embeddings in the forward propagation like QAT does, the gradients of the step size will always be zero as the embeddings are already in low-precision format. 
To overcome the above challenges, we propose adaptive low-precision training (ALPT), that is to learn the low-precision embeddings and the step size alternately as described in Algorithm~\ref{alg:alpt}. 

Specifically, we learn the model parameters and the step size in two separate steps. 
Here we use $b$ as a subscript to denote the embeddings or the step size of the features in a batch of the input data and use $\tilde{Q}(\cdot)$ to denote the quantization function that returns the quantized integers. In the first step, the integer embeddings $\mbw_b^t$ will be de-quantized into floating point values $\mhw_b^t$ for the forward propagation, and then $\mhw_b^t$ will be updated into full-precision weights $\mw_b^{t+1}$.   
In the second step, the model will take $\mw_b^{t+1}$ as input and quantize them with $\md_b^t$ in the forward propagation similar as LSQ.  In this way, we can obtain gradients for the step size. After two steps of optimization, both the model parameters and step size are  optimized. Then, we quantize $\mw_b^{t+1}$ back into integers with $\md_b^{t+1}$ with SR. 
Inspired by LSQ, to ensure convergence, we scale the gradient of the step size and adjust its learning rate to achieve optimal performance. Specifically, we set the scaling factor $g=1/{\sqrt{bdq}} $, where $b$ is the batch size, $d$ is the embedding dimension and ${q}=2^{m-1}-1$. 

\begin{algorithm}[t] 
    \caption{Adaptive low-precision training}\label{alg:alpt}
    \begin{algorithmic}[1]
        \REQUIRE integer embeddings $\mbw_b^t$, step size $\md_b^t$, other parameters $\mw_o^t$ and the loss function $f$. \\
        {\color{blue} {\# Step 1: Update the weights}}
        \STATE {$ \mhw_b^t = \md_b^t \mbw_b^t $;}
        \STATE {$ \mw_b^{t+1} = \mhw_b^t - \eta^t \frac{ \partial f(\mhw_b^t, \mw_o^t) }{ \partial \mhw_b^t }$; }
        \STATE {$ \mw_o^{t+1} = \mw_o^{t} - \eta ^t \frac{ \partial f(\mhw_b^t, \mw_o^t) }{ \partial \mw_o^t }$.}\\
        {\color{blue}\# Step 2: Update the step size}
        \STATE{$ \md_b^{t+1} = \md_b^{t} - \eta ^t \frac{ \partial f(Q_D(\mw_b^{t+1}, \md_b^{t}),\mw_o^{t+1})}{ \partial \md_b^{t} }$; }
        \STATE{$ \mbw_b^{t+1} = \tilde{Q}_S(\mw_b^{t+1}, \md_b^{t+1})$;}
        \ENSURE $\mbw_b^{t+1}, \md_b^{t+1}, \mw_o^{t+1}$.
    \end{algorithmic} 
\end{algorithm}

\section{Experiments}\label{sec:experiment}

\subsection{Evaluation protocol}\label{sec:protocol}
\subsubsection{Dataset}
In this section, we conduct experiments on two real-world datasets: Criteo~\footnote{\url{https://www.kaggle.com/c/criteo-display-adchallenge}} and 
Avazu~\footnote{\url{https://www.kaggle.com/c/avazu-ctr-prediction}}. 
\begin{itemize}
    \item The Criteo dataset consists of 26 categorical feature fields and 13 numerical feature fields. We discretize each numeric value $x$ to $\lfloor \log^2(x)\rfloor$, if $x \textgreater 2$; $x=1$ otherwise. For categorical features, we replace the features that appear less than 10 times with a default "OOV" token.
    \item The Avazu dataset consists of 23 categorical feature fields. We transform the timestamp field into three new fields: hour, weekday, and is\_weekend. Further, we replace the categorical features that appear less than twice with a default "OOV" token.
\end{itemize}
For both datasets, we split them in a ratio of 8:1:1 randomly to get corresponding training, validation, and test sets. Note that the pre-processing rules are similar to~\cite{ctr_benchmark}.

\subsubsection{Settings}\label{sec:settings}
As presented in~\cite{ctr_benchmark}, the performance of different deep CTR models are similar to each other, therefore we choose DCN~\cite{dcn} as the backbone model. The embedding dimension is primarily set to 16 in our experiments. See Appendix~\ref{sec:detailed} for details.
We measure the performance of the ALPT and the baselines in terms of AUC and Logloss, which are two commonly-used metrics for CTR prediction. Note that an increase of 0.001 in AUC is generally considered as a  significant improvement for CTR prediction asks~\cite{wide_deep}. 

To ensure that the compared baselines are sufficiently tuned, we refer to the open benchmark for CTR prediction \cite{ctr_benchmark} to set up our experiments. We use Adam~\cite{kingma2014adam} as the optimizer. The learning rate is set to 0.001 and the maximum number of epochs is 15. We reduce the learning rate tenfold after the 6th and 9th epochs. For regularization, we set the weight decay of embeddings to $5e-8$ and $1e-5$ for Avazu and Criteo,  respectively. In addition, we adpot a dropout of 0.2 on MLP for Criteo. 
For the step size of ALPT, we set its learning rate to $2e-5$ and adopt the same weight decay and learning rate decay as the embeddings. 
All the experiments are run on a single Tesla V100 GPU with Intel Xeon Gold-6154 CPUs. Each experiment is performed at least five times.

\subsubsection{Baselines}
To show the superiority of ALPT, we set four kinds of baselines:
\begin{itemize}
\item \textbf{FP:} Full-precision embeddings without compression. 
\item \textbf{Hashing} and \textbf{Pruning} for embeddings. We implement the hashing method in~\cite{QRhash} and the embedding pruning method in~\cite{DeepLight} according to the instructions in the corresponding papers. 
\cite{QRhash} uses the quotient ($\texttt{id}/r$) and remainder ($\texttt{id}\%r$) to index two embeddings, where $\texttt{id}$ is the feature id and $r$ is the compression ratio. The two embeddings will be multiplied as the final embedding. 
\cite{DeepLight} prunes and retrains the embeddings, where the pruning ratio gradually increases. See Appendix~\ref{sec:detailed} for details.

\item QAT for embeddings. We implement two of the SOTA methods: \textbf{PACT}~\cite{pact} and \textbf{LSQ}~\cite{lsq-EsserMBAM20}. LSQ learns the step size by gradient descent. Similarly, PACT is to learn the clipping value (i.e., the range of quantized weights). Note that we use DR in LSQ and PACT since DR is the common choice in QAT.
\item \textbf{LPT} for embeddings: We implement the vanilla LPT in \cite{baidu_int16}. Following \cite{baidu_int16}, we tune the clipping value among [1, 0.1, 0.01, 0.001]. For LPT and ALPT, we use SR unless we add extra annotations. 
\end{itemize}

\subsection{Overall performance}\label{sec:overall}

\begin{table*}[t]
  \centering
   \caption{{Performance of ALPT and the baselines on Criteo and Avazu.}}
   \resizebox{.99\textwidth}{!}{
    \begin{tabular}{l|ccc|ccc|cc}
    \hline
     & \multicolumn{3}{c}{Avazu} & \multicolumn{3}{|c}{Criteo} & \multicolumn{2}{|c}{Compression ratio}\\
    \hline
     & AUC & Logloss & Epochs $\times$ Time  
     & AUC & Logloss & Epochs $\times$ Time & Training & Inference \\
    \hline
    FP                          & 0.7949($\pm$2e-4)  &  0.37069($\pm$1e-4) & 1 $\times$  6min 
                                & 0.8144($\pm$5e-5)  & 0.43764($\pm$5e-5)  & 9 $\times$  9min &1x &1x \\
                                \hline
    Hashing                     & 0.7928($\pm$3e-4)  & 0.37203($\pm$2e-4)   & 1 $\times$  6min 
                                & 0.8119($\pm$7e-5)  & 0.44130($\pm$6e-5)   & 11 $\times$  9min &2x &2x \\
    Pruning                     & 0.7926($\pm$2e-4)  & 0.37202($\pm$1e-4)   & 1 $\times$  27min 
                                & 0.8123($\pm$5e-5)  & 0.44049($\pm$5e-5)   & 8 $\times$  16min &1x &2x \\
                                \hline
    PACT                        & 0.7948($\pm$2e-4) &  0.37074($\pm$1e-4)  & 1 $\times$  6min 
                                & 0.8144($\pm$8e-5)  & 0.43765($\pm$6e-5)  & 9 $\times$  9min &1x &4x \\
    LSQ                         & 0.7949($\pm$1e-4) &  0.37073($\pm$1e-4)  & 1 $\times$  6min 
                                & 0.8144($\pm$3e-5)  & 0.43764($\pm$3e-5)  & 9 $\times$  9min &1x &4x \\    
                                \hline
    LPT(DR)                     & 0.7654($\pm$4e-4) &  0.38844($\pm$3e-4)  & 15 $\times$  6min 
                                & 0.7966($\pm$1e-4)  & 0.45306($\pm$1e-4)  & 15 $\times$  9min &4x &4x \\    
    LPT(SR)                     & 0.7927($\pm$2e-4) &  0.37205($\pm$1e-4)  & 1 $\times$  6min 
                                & 0.8123($\pm$8e-5)  & 0.43945($\pm$3e-5)  & 9 $\times$  9min &4x &4x \\    
                                \hline
    ALPT(DR)                    & 0.7928($\pm$3e-4) &  0.37216($\pm$2e-4)  & 1 $\times$  7min 
                                & 0.8096($\pm$2e-4)  & 0.44198($\pm$2e-4)  & 6 $\times$  11min &3.2x &3.2x \\
    ALPT(SR)                    & \textbf{0.7951($\pm$2e-4)} &  \textbf{0.37062($\pm$1e-4)}  & 1 $\times$  7min 
                                & \textbf{0.8144($\pm$3e-5)}  & \textbf{0.43763($\pm$3e-5)}  & 9 $\times$  11min &3.2x &3.2x \\
    \hline
  \end{tabular}\label{tab:overall}
}
\end{table*} 

\begin{table*}[!]
  \centering
  \small 
   \caption{Accuracy of different quantization methods with smaller bit widths.}\label{tab:bit}
   \resizebox{.65\textwidth}{!}{
    \begin{tabular}{l|cc|cc|cc|cc}
    \hline
     & \multicolumn{4}{c}{Avazu} & \multicolumn{4}{|c}{Criteo} \\
    \hline
    & \multicolumn{2}{c}{2-bit} & \multicolumn{2}{|c}{4-bit} 
    & \multicolumn{2}{|c}{2-bit} & \multicolumn{2}{|c}{4-bit}  \\
    \hline
     & AUC & Logloss  & AUC & Logloss  & AUC & Logloss & AUC & Logloss \\
    \hline
    PACT                    & 0.7678 &  0.38604  & 0.7925 & 0.37211 
                            & 0.7900 &  0.45945  & \textbf{0.8114} & \textbf{0.44030}  \\
    LSQ                     & \textbf{0.7912} &  \textbf{0.37317}  & \textbf{0.7940} & \textbf{0.37129} 
                            & \textbf{0.8089} &  \textbf{0.44296}  & 0.8105 & 0.44278  \\
    LPT(SR)                 & 0.7829 &  0.37806  & 0.7906 & 0.37342 
                            & 0.8009 &  0.44973  & 0.8079 & 0.44379  \\
    ALPT(SR)                & 0.7851 &  0.37674  & 0.7919 & 0.37260 
                            & 0.8033 &  0.44760  & 0.8111 & 0.44072  \\
    \hline
  \end{tabular}
  }
\end{table*} 

\begin{table*}[!]
  \centering
  \small 
   \caption{Accuracy with larger embedding dimension and more categorical features.}\label{tab:dn}
   \resizebox{.65\textwidth}{!}{
    \begin{tabular}{l|cc|cc|cc|cc}
    \hline
     & \multicolumn{4}{c}{Avazu} & \multicolumn{4}{|c}{Criteo} \\
    \hline
    & \multicolumn{2}{c}{d=32} & \multicolumn{2}{|c}{threshold=1} 
    & \multicolumn{2}{|c}{d=32} & \multicolumn{2}{|c}{threshold=2}  \\
    \hline
     & AUC & Logloss  & AUC & Logloss  & AUC & Logloss & AUC & Logloss \\
    \hline
    FP                      & 0.7951 &  0.37055  & 0.7945 & 0.37099 
                            & 0.8123 &  0.44039  & 0.8125 & 0.43975  \\
    LPT(SR)                 & 0.7934 &  0.37162  & 0.7923 & 0.37234 
                            & 0.8119 &  0.44013  & 0.8115 & 0.44036  \\
    ALPT(SR)                & \textbf{0.7955} &  \textbf{0.37040}  & \textbf{0.7946} & \textbf{0.37098} 
                            & \textbf{0.8126} &  \textbf{0.44000}  & \textbf{0.8126} & \textbf{0.43948}  \\
    \hline
  \end{tabular}
  }
\end{table*}

In this section, we will compare the performance of ALPT and the baselines. Also, we will verify our theoretical analysis about using SR or DR in LPT and ALPT. Note that the bit width of quantization is set to 8 and the compression ratio of hashing and pruning is set to 2$\times$.

As Table \ref{tab:overall} shows, ALPT achieves lossless compression and obtains the best accuracy. 
In contrast, the accuracy degradation of LPT, hashing or pruning is severe and unacceptable. Compared to LPT, ALPT has significant improvement on the prediction accuracy. Compared to the hashing and pruning methods, ALPT enables higher compression ratio and better accuracy at the same time. 
Although LSQ and PACT achieve comparable prediction accuracy, ALPT outperforms them at the training memory usage. Note that we also count the storage of the step size into the embedding size which slightly weakens the compression capability of ALPT. However, as the embedding dimension increases, the effect of the step size on the compression ratio is negligible.

As for the efficiency, we report the training time and inference time of various methods. Specifically, we record the average training time of each epoch on the training set and the average inference time of each step on the validation set. 
The inference time of different methods is similar (i.e., about 150 ms for each step). 
The training time of most methods is similar and ALPT only takes an extra minute on Avazu and two extra minutes on Criteo to learn the step size.

Additionally, as shown in Table \ref{tab:overall}, SR always achieves better performance than DR. Specifically, in LPT, SR has a better accuracy and can converge in fewer epochs, which is consistent with our analysis in Section~\ref{sec:sr-dr}. While ALPT can improve the convergence rate of DR by learning the step size, however, the convergence accuracy is still far from SR.

\subsection{Scalability}\label{sec:scalability}
In this section, to further validate the effectiveness of ALPT, we study the scalability of ALPT from three aspects, that is the bit width of quantization, the embedding dimension and the feature numbers, respectively.

\subsubsection{Smaller bit widths}

For the quantization methods in Table~\ref{tab:overall}, we set the bit width to 4 and 2, respectively. Note that we have tuned the clipping value among [1, 0.1, 0.01, 0.001] for LPT and set the clipping value to 0.1 for 2-bit and 4-bit quantization. In ALPT, we adopt smaller weight decay for the step size (i.e., 0 for Avazu and 1e-6 for Criteo). Since lower bit width has a smaller representation range, the corresponding step size should be larger. 
As shown in Table~\ref{tab:bit}, with different bit widths, the performance of ALPT is also consistently higher than that of LPT, especially in 2-bit. However, the accuracy gap between ALPT and LSQ becomes larger as the bit width decreases.

\subsubsection{Larger embedding dimension}

In the above experiments, the embedding dimension is set to 16. Here, we increase the embedding dimension to 32 and set the bit width to 8. As shown in Table~\ref{tab:dn}, with d=32, ALPT significantly improves the accuracy of LPT and even slightly surpasses the full-precision embeddings.
\subsubsection{More categorical features}
In Section~\ref{sec:settings}, we replace the features that appear less than twice in Avazu or 10 times in Criteo with a default "OOV" token, which determines that Avazu has 4428293 features and Criteo has 1086895 features.  In this section, to obtain datasets with more categorical features, we decrease the threshold, that is from 2 to 1 for Avazu and from 10 to 2 for Criteo. In this way, Criteo has 6780382 features and Avazu has 9449238 features. We conduct experiments with the regenerated datasets to validate the scalability of ALPT on larger datasets. As shown in Table~\ref{tab:dn}, ALPT has always achieved lossless compression, which is a comprehensive proof of its good scalability.

\subsection{Hyper-parameters}
\begin{figure}[!t]
\centering
\subfigure[Avazu] 
{
\begin{minipage}[t]{0.99\linewidth}
\includegraphics[scale=0.35]{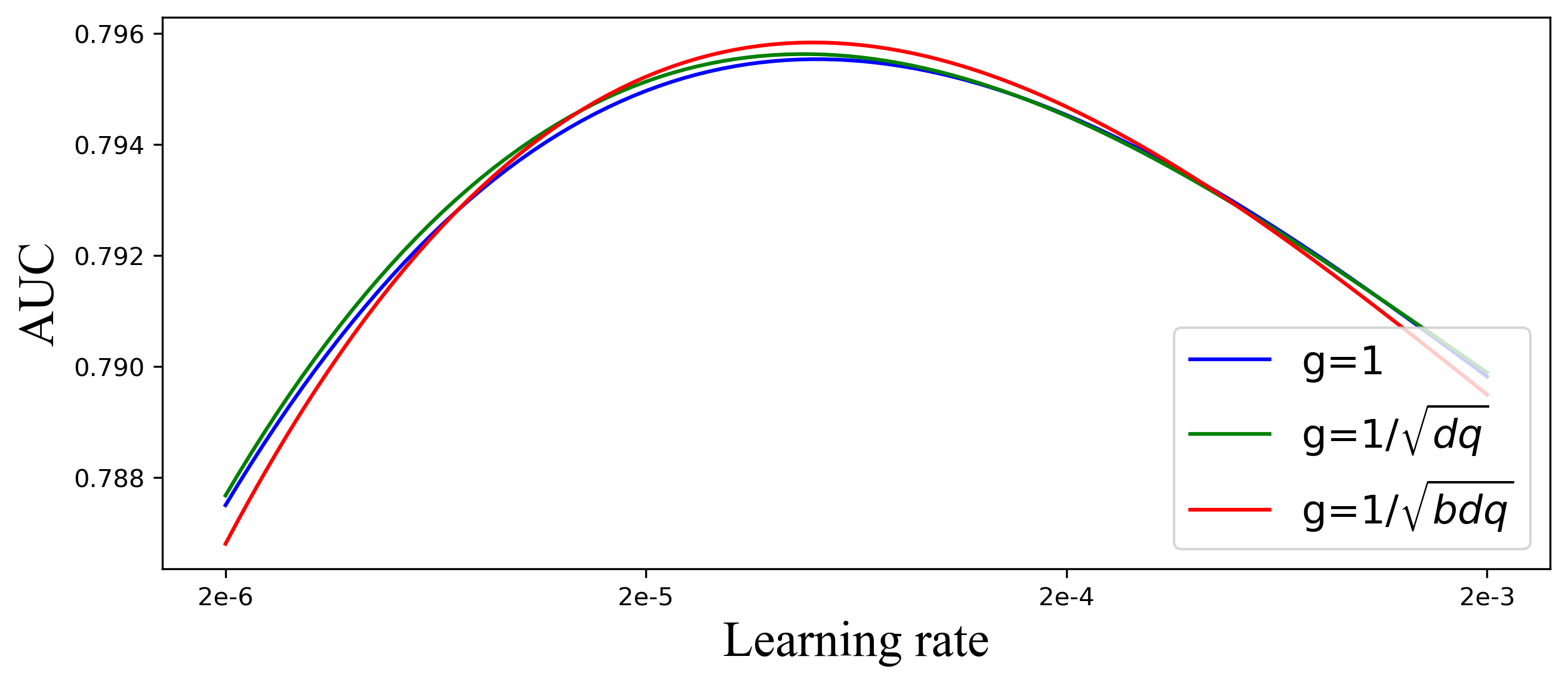}
\end{minipage}
}
\subfigure[Criteo] 
{
\begin{minipage}[t]{0.99\linewidth}
\includegraphics[scale=0.31]{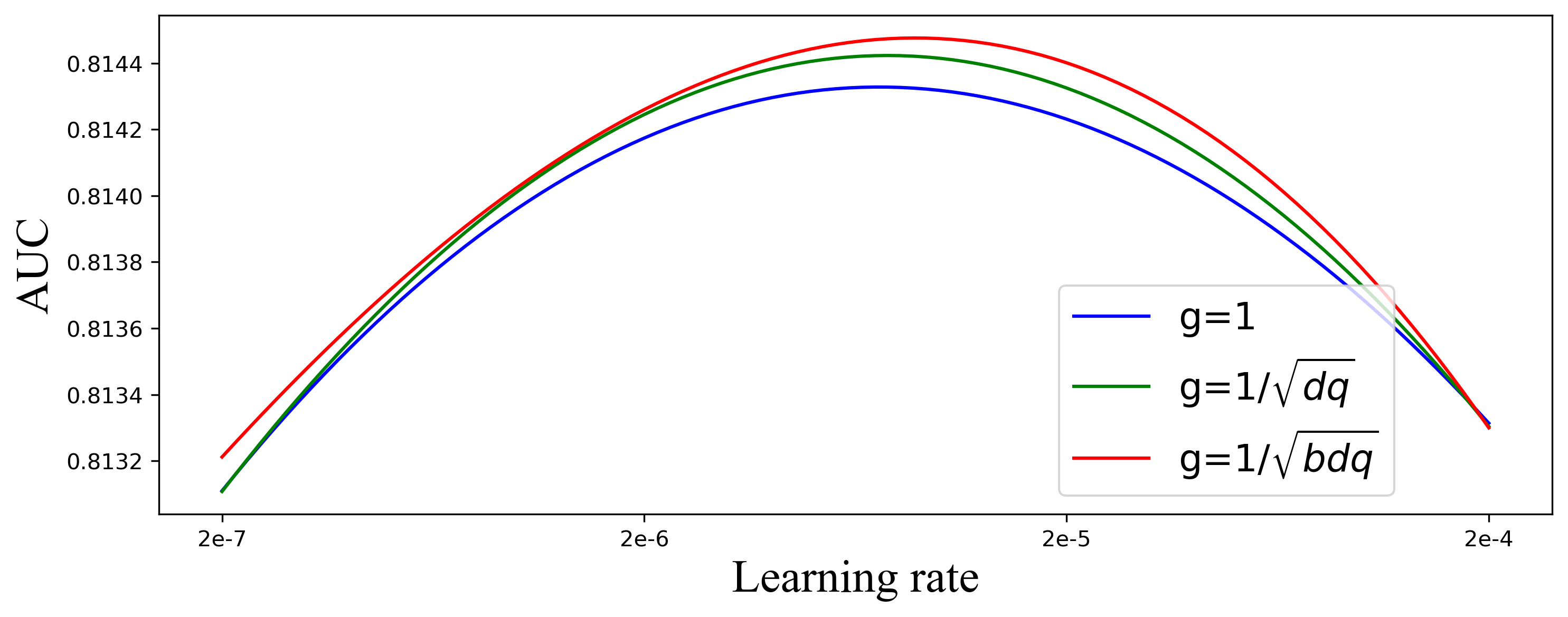}
\end{minipage}
}
\caption{AUC under different learning rates and gradient scaling factors of the step size.} 
\label{fig:lr}
\end{figure}

Inspired by LSQ, we scale the gradient of the step size. The scaling factor $g$ is tuned among $[1, 1/{\sqrt{dq}}, 1/{\sqrt{bdq}}]$. However, we find that the gradient scaling has little influence on accuracy, instead the learning rate has a significant effect.  As Figure \ref{fig:lr} shows, different scaling factors has similar accuracy  given the learning rate. In our analysis, each step size is only responsible for representing the weights in the corresponding embedding, which is easy to fit, so gradient scaling has little effect. In contrast, the weight decay on the step size is more sensitive to the learning rate, which makes the learning rate have a significant impact on the final performance.

\section{{Related work}}\label{sec:related-work}
In this paper, a novel adaptive low-precision training paradigm is proposed for compressing the embedding tables at the training stage. The most related domains are embedding compression and quantization. In this section, we discuss related work in these two domains.

\subsection{Embedding compression}
As we mentioned in Section~\ref{sec:introduction}, research on embedding compression in recommender systems mostly focus on three aspects: NAS-based approaches, embedding pruning and hashing. 
Specifically, NAS-based approaches search a proper dimension for each feature to save memory and improve the prediction accuracy. For example, \cite{NIS} uses a reinforcement learning algorithm to search the optimal embedding dimension for users and items, while \cite{AutoDim} adopts differential architecture search (DARTS) algorithm~\cite{DARTS} to learn the embedding dimension for each feature field. 
Considering that the search processes of these methods are quite time-consuming, 
recent works~\cite{Rule, OptEmbed} search for the optimal embedding dimension with well-designed search strategies. 
Similarly, parameter pruning reduce the number of weights in the embedding tables with an unstructured manner. 
\cite{DeepLight} prunes and retrains the embedding table alternatively so that the mistakenly pruned weights can grow back. 
\cite{PEP} maintains a learnable threshold to prune the embedding weights. The learnable threshold will be updated together with other parameters. 
Different from them, \cite{QRhash} converts an embedding table into two smaller matrices by two hash functions. Further, \cite{doublehash} combine frequency hashing with double hashing for better accuracy. 
Also, recent studies~\cite{TT,TT-KD,STT} have adopted tensor train decomposition to compress the embedding tables and maintain the prediction accuracy through knowledge distillation. 

\subsection{Quantization}
The deep learning community has extensive research and applications on quantization, which can be further divided into two sub-categories, that is quantizing a pre-trained model, i.e., post-training quantization (PTQ) ~\cite{post-aciq,post-embedding,adaround} or training a quantized model from scratch~\cite{lsq-EsserMBAM20,hou2017loss}. 
To train a quantized model from scratch, most of the works follow the quantization-aware training paradigm  \cite{courbariaux2015binaryconnect,rastegari2016xnor}, which quantizes weights in the forward pass and updates the full-precision weights with the gradients of the quantized weights. Recent works have also studied the loss-aware quantization \cite{hou2018loss,hou2019analysis} which explicitly considers the effect of quantization on the loss function. 

In addition to different quantization paradigms, many works also explore the key factors in quantification, such as the rounding functions and  the step size (or the clipping values). For example, \cite{lsq-EsserMBAM20,pact} learn the step size and the clipping values by gradient descent, \cite{hou2018loss} approximates the optimal clipping values by second-order optimization. \cite{adaround} studies the rounding function in PTQ and propose adaptive rounding. Similarly, recent works have explored the strength of stochastic rounding. Note that if we ignore the error caused by clipping function, stochastic weight quantization is equivalent to adding a zero-mean error term (i.e. Gaussian noise) to the weights, which can also be seen as a form of regularization.
\cite{courbariaux2015binaryconnect} studies stochastic weight binarization, \cite{lin2016fixed} considers studies weight ternarization. \cite{hou2019analysis,brain2021luq} consider stochastic gradient quantization.

\section{Conclusion}\label{sec:conclusion}
In this paper, we formulate the low-precision training paradigm for compressing embedding tables at the training stage. 
We provide theoretical analysis on its convergence with stochastic and deterministic weight quantization, which shows that stochastic weight quantization is more suitable for LPT. To reduce accuracy degradation, we further propose adaptive low-precision training (ALPT) to learn the step size and the embeddings alternately. We conduct experiments on two real-world datasets which confirm our analysis while validating the superiority and scalability of ALPT.

\section{Acknowledgments}
This work is supported in part by National Natural Science Foundation of China under grants U1836204, U1936108, 62206102,  and Science and Technology Support Program of Hubei Province under grant 2022BAA046.

\bibliography{aaai23}

\clearpage

\appendix \label{sec:appendix}

\section{Proofs} \label{sec:proof}
\cite{li2017training} already proves Theorem~\ref{thm:stoc_conv}, here we provide proofs for Theorem~\ref{thm:det_conv}. Before that, we first give the quantization error of Eq.(\ref{eq:lpt}) and the following lemmas. 

We denote the quantization error of $w_i^t$ by $r^t_i$. For simplicity, we further define $q=\frac{-\eta^t \nabla f^t(w^t_i)}{\Delta} - \lfloor\frac{-\eta^t \nabla f^t(w^t_i)}{\Delta}\rfloor$, where $q\in[0,1]$. 
With deterministic rounding, $r_i^t$ has the following values: 

\begin{equation}
\begin{aligned}
r^t_i &= Q_D(w^t_i-\eta^t\nabla f^t(w^t_i))-(w^t_i-\eta^t\nabla f^t(w^t_i))\\
&=\Delta*\left\{
\begin{array}{ll}
-q    & \text{if} \quad q\textless 0.5,\\ 
-q+1  & \text{otherwise}.
\end{array}
\right.
\end{aligned}\label{eq:r}
\end{equation}

\begin{lem}
\label{lem:quan_error}
The quantization error of deterministic rounding is bounded as:
\begin{equation}
\CE(||\r^t||^2) \leq \sqrt{d}\Delta\eta^t G.
\end{equation}
\end{lem}

\begin{lem}
\label{lem:det_quan_error}
Given $T_0=\lfloor \frac{2\eta G} {\sqrt{d}\Delta} \rfloor$, the quantization error is also bounded as: 
\begin{equation}
\begin{aligned}
\CE(||\r^t||^2) &\leq \min({\frac{d\Delta^2}{4}}, (\eta^t)^2G^2) \\
&=\left\{
\begin{array}{ll}
{\frac{d\Delta^2}{4}}   & \text{if} \quad t\leq T_0,\\ 
(\eta^t)^2G^2  & \text{otherwise}.
\end{array}
\right.
\end{aligned}
\end{equation}
\end{lem}

\subsection*{Proof of Lemma~\ref{lem:quan_error}}
With deterministic rounding, we have 
\begin{equation}
\begin{aligned}
(r^t_i)^2 &= \Delta ^2 (\min \{q,1-q\})^2\\
&\leq \Delta ^2 \min \{q,1-q\}.
\end{aligned}
\end{equation}

As $\min\{q,1-q\} \leq |\frac{\eta^t\nabla f^t(w_i^t)}{\Delta }|$, the quantization error of deterministic rounding follows:
\begin{equation}
\begin{aligned}
(r^t_i)^2 &\leq \Delta ^2 |\frac{\eta^t\nabla f^t(w_i^t)}{\Delta }| 
\leq \Delta \eta^t |\nabla f^t(w_i^t)|.
\end{aligned}
\end{equation}

Summing over the index $i$ and taking expectation:
\begin{equation}
\begin{aligned}
\CE(||\r^t||^2) &\leq \Delta \eta^t \CE|| \nabla f^t(\mw^t)||_1 \\ 
&\leq \sqrt{d}\Delta \eta^t \CE||\nabla f^t(\mw^t)||.
\end{aligned}
\end{equation}

As $(\CE||\nabla f^t(\mw^t)||)^2 \leq \CE||\nabla f^t(\mw^t)||^2 \leq G^2$, we have:

\begin{equation}
\begin{aligned}
\CE(||\r^t||^2) \leq \sqrt{d}\Delta \eta^t G.
\end{aligned}
\end{equation}

\subsection*{Proof of Lemma~\ref{lem:det_quan_error}}
From Eq(\ref{eq:r}), we have:
\begin{equation}
\begin{aligned}
(r^t_i)^2 &= \Delta ^2 (\min \{q,1-q\})^2\\
&\leq  \Delta ^2 (\frac{1}{2})^2 = \frac{\Delta^2}{4}.
\end{aligned}\label{eq:r1}
\end{equation}

Also, as $\min\{q,1-q\} \leq |\frac{\eta^t\nabla f^t(w_i^t)}{\Delta }|$, we have
\begin{equation}
\begin{aligned}
(r^t_i)^2 &= \Delta ^2 (\min \{q,1-q\})^2\\
&\leq  \Delta ^2 |\frac{\eta^t\nabla f^t(w_i^t)}{\Delta }|^2 \\
&= (\eta^t)^2 |\nabla f^t(w_i^t)|^2.
\end{aligned}\label{eq:r2}
\end{equation}

Summing over the index $i$ and taking expectation with Eq.(\ref{eq:r1}) and Eq.(\ref{eq:r2}), we have
\begin{equation}
\begin{aligned}
\CE(||\r^t||^2) &\leq \min({\frac{d\Delta^2}{4}}, (\eta^t)^2G^2).
\end{aligned}
\end{equation}

Considering that $\eta^t$ decays like $\frac{\eta}{\sqrt{t}}$, we have $T_0=\lfloor \frac{2\eta G} {\sqrt{d}\Delta} \rfloor$ makes:
\begin{equation}
\begin{aligned}
\min({\frac{d\Delta^2}{4}}, (\eta^t)^2G^2) =\left\{
\begin{array}{ll}
{\frac{d\Delta^2}{4}}   & \text{if} \quad t\leq T_0,\\ 
(\eta^t)^2G^2  & \text{otherwise}.
\end{array}
\right.
\end{aligned}
\end{equation}

\subsection*{Proof of Theorem~\ref{thm:det_conv}}
The weights are updated in LPT with DR as:
\begin{equation}
\begin{aligned}
\mw^{t+1} = Q_D(\mw^t - \eta^t \nabla f^t(\mw^t))= \mw^t - \eta^t \nabla f^t(\mw^t) + \r^t.
\end{aligned}
\end{equation}

Hence, we have
\begin{equation}
\begin{aligned}
&||\mw^{t+1}-\mw^*||^2 \\
=& || \mw^t - \eta^t \nabla f^t(\mw^t) + \r^t -\mw^*||^2 \\
=&|| \mw^t -\mw^*||^2 + ||\eta^t \nabla f^t(\mw^t) - \r^t\\
& - 2\langle \mw^t -\mw^*, \eta^t \nabla f^t(\mw^t) - \r^t\rangle \\
=&|| \mw^t -\mw^*||^2 + ||\eta^t \nabla f^t(\mw^t)||^2  \\
& +||\r^t||^2 - 2\langle \mw^t - \mw^*, \eta^t \nabla f^t(\mw^t)\rangle \\ 
& + 2\langle \mw^t -\mw^*, \r^t\rangle + 2\langle \eta^t \nabla f^t(\mw^t), - \r^t\rangle.
\end{aligned}
\end{equation}

After rearranging, 
\begin{equation}\label{eq:product}
\begin{aligned}
&\langle \mw^t - \mw^*, \nabla f^t(\mw^t)\rangle \\
=&\frac{(||\mw^t -\mw^*||^2 - || \mw^{t+1} -\mw^*||^2)}{2\eta^t} \\
& + \frac{\eta^t ||\nabla f^t(\mw^t)||^2 }{2}
+\frac{||\r^t||^2 }{2\eta^t} \\ 
&+ \frac{\langle \mw^t -\mw^*, \r^t\rangle}{\eta^t}
+ \langle \nabla f^t(\mw^t), - \r^t\rangle
\end{aligned}
\end{equation}

Taking expectation with $\CE(\nabla  f^t(\mw^t))=\nabla  F(\mw^t)$, we have 

\begin{equation}\label{eq:product2}
\begin{aligned}
& \CE[\langle \mw^t - \mw^*, \nabla F(\mw^t)\rangle] \\
\leq& \frac{(\CE||\mw^t -\mw^*||^2 - \CE|| \mw^{t+1} -\mw^*||^2)}{2\eta^t} \\
&+\frac{\eta^t \CE||\nabla f^t(\mw^t)||^2 }{2}
+\frac{\CE||\r^t||^2 }{2\eta^t} \\ 
& + \frac{\CE\langle  \mw^t -\mw^*, \r^t\rangle}{\eta^t} \\ 
& + \CE\langle \nabla f^t(\mw^t), - \r^t\rangle
\end{aligned}
\end{equation}

With Lemma~\ref{lem:quan_error} and the bounded gradients assumption $\CE||\nabla  f^t(\mw^t)||^2\leq G^2$, we have 

\begin{equation}\label{eq:product2}
\begin{aligned}
& \CE[\langle \mw^t - \mw^*, \nabla F(\mw^t)\rangle] \\
& \leq \frac{(\CE||\mw^t -\mw^*||^2 - \CE|| \mw^{t+1} -\mw^*||^2)}{2\eta^t} \\ 
& \quad 
+ \frac{\eta^t G^2}{2} 
+\frac{\sqrt{d}\Delta G}{2}   + \frac{\CE\langle  \mw^t -\mw^*, \r^t\rangle}{\eta^t} \\
& \quad
+ \CE\langle   \nabla f^t(\mw^t), - \r^t\rangle
\end{aligned}
\end{equation}

Based on Cauchy–Schwarz inequality and Lemma~\ref{lem:det_quan_error}, we have 

\begin{equation}\label{eq:cs2}
\begin{aligned}
\CE\langle  \nabla f^t(\mw^t), - \r^t\rangle
& \leq |\CE\langle  \nabla f^t(\mw^t), - \r^t\rangle| \\
& \leq \sqrt{\CE||\nabla f^t(\mw^t)||^2 \CE||\r^t||^2} \\
& \leq \eta^t G^2,
\end{aligned}
\end{equation}

\begin{equation}\label{eq:cs1}
\begin{aligned}
\CE\langle  \mw^t -\mw^*, \r^t\rangle 
& \leq |\CE\langle  \mw^t -\mw^*, \r^t\rangle | \\ 
& \leq \sqrt{\CE||\mw^t -\mw^*||^2  \CE ||\r^t||^2} \\ 
& \leq \left\{
\begin{array}{ll}
{\frac{\sqrt{d}D\Delta}{2}}   & \text{if} \quad t\leq T_0,\\
\eta^tDG  & \text{otherwise}.
\end{array}
\right.
\end{aligned}
\end{equation}

With Eq.(\ref{eq:cs1}) and Eq.(\ref{eq:cs2}), we can have from Eq.(\ref{eq:product2})

\begin{equation}
\begin{aligned}
&\CE[\langle \mw^t - \mw^*, \nabla F(\mw^t)\rangle]  \\
 \leq &\frac{(\CE||\mw^t -\mw^*||^2  - \CE|| \mw^{t+1} -\mw^*||^2 )}{2\eta^t}  
+ \frac{3\eta^t G^2}{2} \\ &
+\frac{\sqrt{d}\Delta G}{2} + \min(\frac{\sqrt{d}D\Delta}{2\eta^t}, DG)
\end{aligned}
\end{equation}

Since $F(x)$ is convex, $F(\mw^t)-F(\mw^*) \leq \langle \mw^t - \mw^*, \nabla F(\mw^t)\rangle$, we have

\begin{equation} \label{eq: repeat_start}
\begin{aligned}
&\CE[F(\mw^t)-F(\mw^*)] \\
\leq &\frac{(\CE||\mw^t -\mw^*||^2  - \CE|| \mw^{t+1} -\mw^*||^2 )}{2\eta^t}  
+ \frac{3\eta^t G^2}{2} \\ &
+\frac{\sqrt{d}\Delta G}{2} + \min(\frac{\sqrt{d}D\Delta}{2\eta^t}, DG)
\end{aligned}
\end{equation}

Assuming $T_0\textless T$ and accumulating from $t=1$ to $T$,

\begin{equation}
\begin{aligned}
& \sum_{t=1}^{T} \CE[F(\mw^t)-F(\mw^*)] \\
& \leq \sum_{t=1}^{T}\frac{(\CE||\mw^t -\mw^*||^2  - \CE|| \mw^{t+1} -\mw^*||^2 )}{2\eta^t} \\ 
& + \sum_{t=1}^{T}\frac{3\eta^t G^2}{2} + \frac{T\sqrt{d}\Delta G}{2} + \sum_{t=1}^{T}\min(\frac{\sqrt{d}D\Delta}{2\eta^t}, DG) \\ 
& = \frac{1}{2\eta_1}\CE|| w_{1} -\mw^*||^2 
+ \sum_{t=2}^{T}(\frac{1}{2\eta^t}-\frac{1}{2\eta_{t-1}})\CE||\mw^t -\mw^*||^2 \\
& + \sum_{t=1}^{T}\frac{3\eta^t G^2}{2}
+ \frac{T\sqrt{d}\Delta G}{2} + \sum_{t=1}^{T_0} \frac{\sqrt{d}D\Delta}{2\eta^t} + \sum_{t=T_0}^{T} DG
\end{aligned}
\end{equation}

Applying $\CE||\mw^t -\mw^*||^2  \leq D^2$ and $\sum_{t=1}^{T} \eta^t \leq 2\eta\sqrt{T}$, we have 

\begin{equation}
\begin{aligned}
&\sum_{t=1}^{T} \CE[F(\mw^t)-F(\mw^*)]  \leq \frac{\sqrt{T}D^2}{2\eta}
+  3\eta\sqrt{T} G^2  \\ &
+ \frac{T\sqrt{d}\Delta G}{2} + \frac{D\Delta\sum_{t=1}^{T_0} \sqrt{t}}{2\eta} + \sum_{t=T_0+1}^{T} DG
\end{aligned}
\end{equation}

As $\bar{\mw}=\frac{1}{T}\sum_{t=1}^T\mw^t$, with Jensen's inequality, we have

\begin{equation}
\begin{aligned}
\CE[F(\bar{\mw}^t)-F(\mw^*)] \leq \frac{1}{T}\sum_{t=1}^{T}\CE[F(\mw^t)-F(\mw^*)]
\end{aligned}
\end{equation}

Eventually, we have
\begin{equation}
\begin{aligned}
\CE[F(\bar{\mw}^t)-F(\mw^*)] & \leq \frac{D^2}{2\eta\sqrt{T}}
+ \frac{3\eta G^2}{\sqrt{T}}
+ \frac{\sqrt{d}\Delta G}{2}
+ DG
\\
&\leq  \frac{D^2}{2\eta \sqrt{T}}
+ \frac{3\eta  G^2}{\sqrt{T}} + \frac{\sqrt{d}D\Delta\sum_{t=1}^{T_0} \sqrt{t}}{2\eta T} \\
&+ \frac{\sqrt{d}\Delta G}{2} + \frac{\sum_{t=T_0+1}^{T}DG}{T}. 
\end{aligned}
\end{equation}

\section{Detailed experimental settings}\label{sec:detailed}
\subsection{Settings}
Here, we give the detailed architecture of DCN used for Avazu and Criteo. DCN consists of two parts: cross network and deep network, which are used to learn low-dimensional and high-dimensional feature interactions, respectively. The depth of both networks is the same. In avazu, the depth of both parts is set to 3 and the width of the deep network is set to 1024, 512, 256. In criteo, the depth of both parts is set to 5 and the width of each layer in the deep network is 1000.

\subsection{Baselines}
Here, we introduce the hashing and pruning methods in our experiments.
\cite{QRhash} divide the embedding table into two smaller embedding tables, $\mathbf{E_1}\in \mathbb{R}^{r\times d}$ and $\mathbf{E_2}\in \mathbb{R}^{\frac{n}{r}\times d}$. Assuming the id of a feature is \texttt{id}, the remainder ($\texttt{id}\%r$) and the quotient ($\texttt{id}/r$) will be used as indexes to get an embedding in $E_1$ and $E_2$, respectively. The two embeddings then will be multiplied as the final embedding. In Section~\ref{sec:overall}, $r$ is set to 2. \cite{DeepLight} will first train the embeddings for a good initialization. It then prunes and retrains the embeddings, where the pruning ratio gradually increases. After the initialization, the pruning ratio can be calculated by $R_x(1-D^{k/U})$, where $R_x$ is the target sparse rate, $k$ is current step, $D$ and $U$ are the damping ratios. In Section~\ref{sec:overall}, $R_x, D, U$ are set to 0.5, 0.99, 3000, respectively.

\end{document}